\documentclass{llncs}
\usepackage{llncsdoc}

\usepackage{graphicx} 
\usepackage{subfigure} 
\usepackage{wrapfig}

\usepackage{psfrag}
\usepackage{amsbsy}
\usepackage[intlimits]{amsmath} 
\usepackage{bbm,amsfonts}

\begin{document}
\markboth{\LaTeXe{} Class for Lecture Notes in Computer
Science}{\LaTeXe{} Class for Lecture Notes in Computer Science}

\pagestyle{empty} 

\title{Gaussian processes for sample efficient reinforcement learning with RMAX-like exploration}

\author{Tobias Jung\inst{1} \and Peter Stone\inst{1}\\
{\tt \{tjung,pstone\}@cs.utexas.edu}}

\institute{Department of Computer Science \\ University of Texas at Austin}

\maketitle

\newcommand{\argmax}{\mathop{\mathrm{argmax}}}
\newcommand{\diag}{\mathop{\mathrm{diag}}}
\newcommand{\norm}[1]{\left\|{#1}\right\|}

\begin{abstract}
We present an implementation of model-based online reinforcement learning (RL) for continuous
domains with deterministic transitions that is specifically designed to achieve low sample complexity. 
To achieve low sample complexity, since the environment is unknown, an agent must intelligently
balance exploration and exploitation, and must be able to rapidly generalize from observations.
While in the past a number of related sample efficient RL algorithms have been proposed,
to allow theoretical analysis, mainly model-learners with weak generalization capabilities 
were considered. 
Here, we separate function approximation in the model learner (which does require samples) 
from the interpolation in the planner (which does not require samples). For model-learning 
we apply Gaussian processes regression (GP) which is able to automatically adjust itself 
to the complexity of the problem (via Bayesian hyperparameter selection) and, in practice, often 
able to learn a highly accurate model from very little data. In addition, a GP provides 
a natural way to determine the uncertainty of its predictions, which allows us to implement the
``optimism in the face of uncertainty'' principle used to efficiently control exploration. 
Our method is evaluated on four common benchmark domains.   
\end{abstract} 

\section{Introduction}
In reinforcement learning (RL), an agent interacts with an environment and 
attempts to choose its actions such that an externally defined performance measure, the 
accumulated per-step reward, is maximized over time.
One defining characteristic of RL is that the environment is {\em unknown} and that the 
agent has to {\em learn} how to act directly from experience. 
In practical applications, e.g., in robotics, obtaining this experience means having a physical 
system interact with the physical environment in real time. Therefore, RL methods that are able 
to learn quickly and minimize the amount of time the robot needs to interact with 
the environment until good or optimal behavior is learned, are highly desirable.   

In this paper we are interested in online RL for tasks with continuous state spaces
and smooth transition dynamics that are typical for robotic control domains. Our primary 
goal is to have an algorithm which keeps sample complexity as low as possible.

\subsection{Overview of the contribution}
To maximize sample efficiency, we consider online RL that is {\em model-based} in the spirit of RMAX \cite{RMAX-JMLR02},
but extended to continuous state spaces similar to \cite{ARL2010,jong07,davies97multidimensional}. 
As in RMAX and related methods, our algorithm, GP-RMAX, consists of two parts: a {\em model-learner} and a {\em planner}. 
The model-learner estimates the dynamics of the environment from the sample transitions the
agent experiences while interacting with the environment. The planner is used to find the best possible action, 
given the current model. As the predictions of the model-learner become increasingly more accurate, 
the actions derived become increasingly closer to optimal. To control the amount of exploration, 
the ``optimism in the face of uncertainty'' principle is employed which makes the agent visit unexplored 
states first. 
In our algorithm, the model-learner is implemented by Gaussian process (GP) regression; being non-parametric, GPs give 
us enhanced modeling flexibility. GPs allow Bayesian model selection and automatic relevance determination. 
In addition, GPs provide a natural way to determine the uncertainty of predictions, which allows us to implement the
``optimism in the face of uncertainty'' exploration of RMAX in a principled way. 
The planner uses the estimated transition function (as estimated by the model) to solve the 
Bellman equation via value iteration on a uniform grid.\footnote{While certainly more advanced methods 
exist, e.g., \cite{Gruene97,Munos-ML02}, for our purpose here, a uniform grid is sufficient as proof of concept.}

The key point of our algorithm is that we separate the steps estimating a function from samples in 
the model-learner from solving the Bellman equation in the planner. The rationale behind this is that, 
if the transition function is relatively simple, it can be estimated accurately from only few sample 
transitions. On the other hand, the optimal value function, due to the inclusion of the max operator, often 
is a complex function with sharp discontinuities.
Solving the Bellman equation, however, does not require actual ``samples''; instead, we must only be able to 
evaluate the Bellman operator in arbitrary points of the state space. This way, when the transition function 
can be learned from only a few samples, large gains in sample efficiency are possible. Competing model-free methods, 
such as fitted Q-iteration \cite{Riedmiller05,Ernst05,nouri08multiresolution} or policy iteration based LSPI/LSTD/LSPE 
\cite{Lagoudakis03,busoniu10onlinelspi,li09LSPIonlineexploration,mein_keepaway2007}, do not have this 
advantage, as they need the actual sample transitions to estimate and represent the value function.

Conceptually, our approach is closely related to Fitted R-MAX, which was proposed in \cite{jong07}
and uses an instance-based approach in the model-learner, and related work in \cite{davies97multidimensional,ARL2010},
which uses grid-based interpolation in the model-learner. The primary contribution of this paper is 
to use GPs instead. Doing this means we are willing to trade off theoretical analysis with 
practical performance. For example, unlike the recent ARL \cite{ARL2010}, for which PAC-style performance bounds 
could be derived (because of its grid-based implementation of model-learning), a GP is much better able to  
handle generalization and as a consequence can achieve much lower sample complexity.

\subsection{Assumptions and limitations}
Our approach makes the following assumptions (most of which are also made in related work,
even if it is not always explicitly stated):

\begin{itemize}
\item {\em Low dimensionality of the state space.} With a uniform grid, the number of grid points 
for solving the Bellman equation scales exponentially with the dimensionality. While
more advanced methods, such as sparse grids or adaptive grids, may allow us to somewhat reduce this
exponential increase, at the end they do not break the curse of dimensionality. 
Alternatively, one can use nonlinear function approximation; however, despite some encouraging results, 
it is unclear as to whether this approach would really do any better in general applications. 
Today, breaking the curse of dimensionality is still an open research problem. 

\item {\em Discrete actions.} While continuous actions may be discretized, in practice, for higher 
dimensional action spaces this becomes infeasible.

\item {\em Smooth transition function.} Performing an action from states that are ``close'' must lead to 
successor states that are ``close''. (Otherwise both the generalization in the model 
learner and the interpolation in the value function approximation would not work).

\item {\em Deterministic transitions.} This is not a fundamental requirement of our approach, since  
GPs can also learn noisy functions (either due to observation noise or random disturbances with small 
magnitude), and the Bellman operator can be evaluated in the resulting predictive distribution. 
Rather it is one taken for convenience.

\item {\em Known reward function.} Assuming that the reward function is known and only the
transition function needs to be learned is what is different from most comparable work. 
While it is not a fundamental requirement of our approach (since we could learn 
the reward function as well), it is an assumption that we think is well justified: for one, 
reward is the performance criterion and specifies the goal. For the type of control problems 
we consider here, reward is always externally defined and never something that is ``generated'' 
from within the environment. Two, reward sometimes is a discontinuous function, e.g., +1 at the goal 
state and 0 elsewhere. Which makes it not very amenable for function approximation. 
\end{itemize}

\section{Background: Planning when the model is exact}
\label{sec:numerical solution}
\begin{figure}[tb]
\psfrag{x00}{\scriptsize $\xi_{00}$}
\psfrag{x10}{\scriptsize $\xi_{10}$}
\psfrag{x11}{\scriptsize $\xi_{11}$}
\psfrag{x01}{\scriptsize $\xi_{01}$}
\psfrag{qza=}{\hspace{-0.24cm}\scriptsize $Q(z,a')=?$}
\psfrag{z=fxa}{\hspace*{-0.35cm}\scriptsize $z:=f(\xi_i,a)$}
\psfrag{xi}{\scriptsize $\xi_i$}
\psfrag{transitionf}{\scriptsize transition $f$}

\psfrag{dx}{\scriptsize $d_x$}
\psfrag{dy}{\scriptsize $d_y$}
\psfrag{hx}{\scriptsize $h_x$}
\psfrag{hy}{\scriptsize $h_y$}
\psfrag{z=(xy)}{\scriptsize $z=(x,y)$}
\psfrag{x0}{\scriptsize $x_0$}
\psfrag{x1}{\scriptsize $x_1$}

	\centering
		\includegraphics[width=0.26\textwidth]{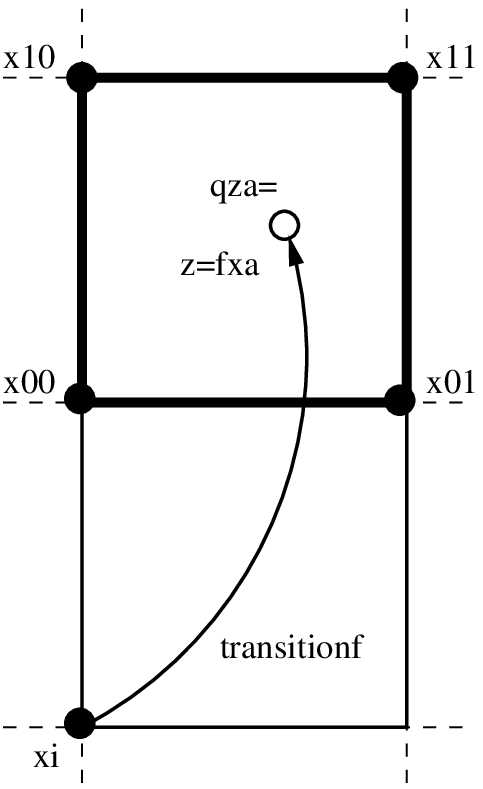} \hspace*{1cm}
		\psfrag{x00}{\scriptsize $(\xi_{00},q_{00}^{a'})$}
        \psfrag{x10}{\scriptsize $(\xi_{10},q_{10}^{a'})$}
        \psfrag{x11}{\scriptsize $(\xi_{11},q_{11}^{a'})$}
        \psfrag{x01}{\scriptsize $(\xi_{01},q_{01}^{a'})$}
		\includegraphics[width=0.36\textwidth]{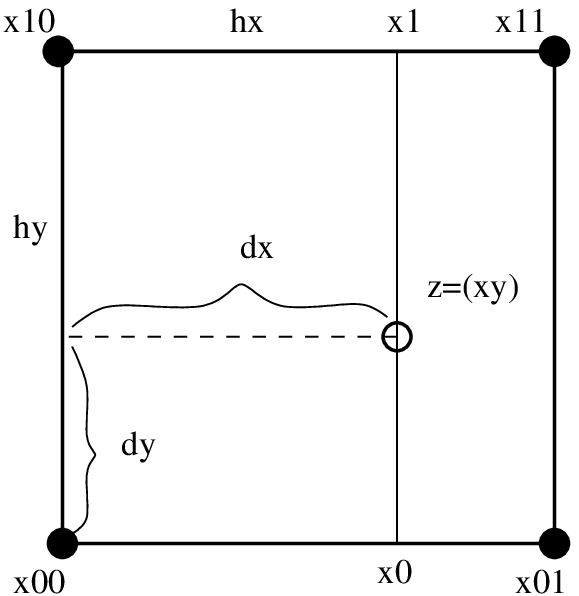}
	\caption{Bilinear interpolation to determine $Q(f(\xi_i,a),a')$ in $\mathbbm R^2$.}
	\label{fig:1}
\end{figure}

Consider the reinforcement learning problem for MDPs with continuous state space, finite action space, 
discounted reward criterion and {\em deterministic} dynamics \cite{sutton98introduction}. In this section 
we assume that dynamics and rewards are available to the learning agent. Let state space $\mathcal X$ be a hyperrectangle in 
$\mathbbm R^d$ (this assumption is justified if, for example, the system is a motor control task), $\mathcal A$ be the
finite action space (assuming continuous controls are discretized), $x_{t+1}=f(x_t,a_t)$ be the transition
function (assuming that continuous time problems are discretized in time), and $r(x,a)$ be the reward 
function.
For the following theoretical argument we require that both transition and reward function are Lipschitz continuous 
in the actions; i.e., there exist constants $L_f,L_r$ such that 
$\norm{f(x,a)-f(x',a)}\le L_f \norm{x-x'}$, and $|r(x,a)-r(x',a)|\le L_r \norm{x-x'}$, $\forall x,x'\in 
\mathcal X, a \in \mathcal A$. In addition, we assume that the reward is bounded, $|r(x,a)|\le R_\text{MAX}$, 
$\forall x,a$. Note that in practice, while the first condition, continuity in the transition function, is 
usually fulfilled for domains derived from physical systems, the second condition, continuity in the rewards, is often violated
(e.g. in the mountain car domain, reward is $0$ in the goal and $-1$ everywhere else). Despite that we find that in many
of these cases the outlined procedure may still work well enough.
 
For any state $x$, we are interested in determining a sequence of actions $a_0,a_1,a_2,\ldots$ such that
the accumulated reward is maximized,
\begin{equation*}
V^*(x):=\max_{a_0,a_1,\ldots} \ \Bigl\{ \sum_{t=0}^\infty \gamma^t r(x_t,a_t) \ | \ x_0=x, x_{t+1}=f(x_t,a_t)\Bigr\},
\end{equation*} 
where $0<\gamma<1$. Using the Q-notation, where $Q^*(x,a):=r(x,a)+\gamma V^*(f(x,a))$, the optimal
decision policy $\pi^*$ is found by first solving the Bellman equation in the unknown function $Q$,
\begin{equation}
Q(x,a)=r(x,a)+\gamma \max_{a'} \, Q(f(x,a),a') \qquad \forall x\in \mathcal X, a\in \mathcal A
\label{eq:1}
\end{equation}
to yield $Q^*$, and then choosing the action with the highest Q-value,
\begin{equation*}
\pi^*(x)=\argmax_{a'} \, Q^*(x,a').
\end{equation*} 
The Bellman operator $T$ related to \eqref{eq:1} is defined by
\begin{equation}
\bigl(TQ\bigr)(x,a):=r(x,a)+\gamma \max_{a'} \, Q(f(x,a),a').
\label{eq:2}
\end{equation}
It is well known that $T$ is a contraction and $Q^*$ the unique bounded solution to the fixed point
problem $Q(x,a)=\bigl(TQ\bigr)(x,a)$, $\forall x,a$.

In order to solve the infinite dimensional problem in \eqref{eq:1} numerically, we have to reduce it to
a finite dimensional problem. This is done by introducing a discretization $\Gamma$ of $\mathcal X$ into
a finite number of elements, applying the Bellman operator to only the nodes and interpolating in between.

In the following we will consider a uniform grid $\Gamma_h$ with $N$ vertices $\xi_i$ and $d$-dimensional 
tensor B-spline interpolation of order $1$. The solution of \eqref{eq:1} is then obtained in the space
of piecewise affine functions. 
\newcommand{\Gh}{{\Gamma_h}}
\newcommand{\QGh}{Q^\Gh}
\newcommand{\TGh}{T^\Gh}

For a fixed action $a'$, the value $\QGh(z,a')$ of any state $z$ with respect to grid $\Gamma_h$ 
can be written as a convex combination of the vertices $\xi_j$ of the grid cell enclosing $z$ with coefficients 
$w_{ij}$ (see Figure~\ref{fig:1}a). 
For example, consider the $2$-dimensional case (bilinear interpolation) in Figure~\ref{fig:1}b. Let $z=(x,y)\in\mathbbm R^2$. 
To determine $\QGh(z,a')$, we find the four vertices $\xi_{00},\xi_{01},\xi_{10},\xi_{11}\in \mathbbm R^2$ of the enclosing cell
with known function values $q_{00}^{a'}:=\QGh(\xi_{00},a'),\ldots$ etc. We then perform two linear interpolations along
the $x$-coordinate (order invariant) in the auxilary points $x_0,x_1$ to obtain
\begin{eqnarray*}
\QGh(x_0,a') &=&(1-\lambda_0)q_{00}^{a'}+\lambda_0 q_{01}^{a'} \\
\QGh(x_1,a') &=&(1-\lambda_0)q_{10}^{a'}+\lambda_0 q_{11}^{a'}
\end{eqnarray*} 
where $\lambda_0:=d_x/h_x$ (see Figure~\ref{fig:1}b for a definition of $d_x,h_x,x_0,x_1$). We then perform another linear 
interpolation in $x_0,x_1$ along the $y$-coordinate to obtain
\begin{equation}
\QGh(z,a')=(1-\lambda_1)(1-\lambda_0)q_{00}^{a'}
+(1-\lambda_1)\lambda_0q_{01}^{a'}
+\lambda_1(1-\lambda_0)q_{10}^{a'}
+\lambda_1 \lambda_0 q_{11}^{a'}
\label{eq:3}
\end{equation}
where $\lambda_1:=d_y/h_y$. Weights $w_{ij}$ now correspond to the coefficients in \eqref{eq:3}. An analogous
procedure applies to higher dimensions.

Let $Q^{a'}$ be the $N \times 1$ vector with entries $[Q^{a'}]_i=\QGh(\xi_i,a')$. Let $z_1^a,\ldots,z_N^a \in \mathbb R^d$
denote the successor state we obtain when  we apply the transition function $f$ to vertices $\xi_i$ using action $a$, i.e., 
$z_i^a:=f(\xi_i,a)$. Let $[w_i^a]_j=w_{ij}^a$ denote the $1 \times N$ vector of coefficients for $z_i^a$ from \eqref{eq:3}. 
The Q-value of $z_i^a$ for any action $a'$ with respect to grid $\Gamma_h$ can thus be written as 
$\QGh(z_i^a,a')=\sum_{j=1}^N [w_i^a]_j [Q^{a'}]_j$. 
Let $W^a$ with rows $[w_i^a]$ be the $N \times N$ matrix of all coefficients. (Note that this matrix is sparse: 
each row contains only $2^d$ nonzero entries).

Let $R^a$ be the $N\times 1$ vector of associated rewards, $[R^a]_i:=r(\xi_i,a)$. Now we can use \eqref{eq:2} to 
obtain a fixed point equation in the vertices of the grid $\Gamma_h$,
\begin{equation}
\QGh(\xi_i,a)=\bigl(\TGh \QGh\bigr)(\xi_i,a) \quad i=1,\ldots,N, \ a=1,\ldots, |\mathcal A|,
\label{eq:4}
\end{equation}
where
\begin{equation*}
 \bigl(\TGh \QGh\bigr)(\xi_i,a):=r(\xi_i,a)+\gamma \max_{a'} \, \QGh(f(\xi_i,a),a').
\end{equation*}
Slightly abusing the notation, we can write this more compactly in terms of matrices and vectors,
\begin{equation}
\TGh \QGh := R^a + \gamma \max_{a'} \, \bigl\{W^a Q^{a'}\bigr\} \quad \forall a.
\label{eq:5}
\end{equation}
The Q-function is now represented by $|\mathcal A|$ $N$-dimensional vectors $Q^{a'}$, each containing the values
for the vertices $\xi_i$. The discretized Bellman operator $\TGh$ is a contraction in $\mathbbm R^d \times \mathcal A$
and therefore has a unique fixed point $Q^* \in \mathbbm R^d \times \mathcal A$.
Let function $Q^{*,\Gh}:(\mathbbm R^d\times \mathcal A) \rightarrow \mathbbm R$ be the Q-function obtained by
linear interpolation of vector $Q^*$ along states. The function $Q^{*,\Gh}$ can now be used to determine (approximately) 
optimal control actions: for any state $x \in \mathcal X$, we simply determine
\begin{equation*}
\pi^{*,\Gh}(x)=\argmax_{a'} \, Q^{*,\Gh}(x,a').
\end{equation*} 
In order to estimate how well function $Q^{*,\Gh}$ approximates the true $Q^*$, {\em a posteriori} estimates can be
defined that are based on local errors, i.e. the maximum of residual in each grid cell. The local error in a grid cell
in turn depends on the granularity of the grid, $h$, and the modulus of continuity $L_f, L_g$ (e.g., see 
\cite{Gruene97,Munos-ML02} for details).

\section{Our algorithm: GP-RMAX}
%
\begin{figure}[t!]
\begin{center}
\psfrag{xtt}{\tiny$x_{t+1}=f(x_t,a_t)$}
\psfrag{xt}{\tiny$x_{t}$}
\psfrag{at}{\tiny$a_{t}$}
\psfrag{triplet}{\tiny\hspace{-0.2cm}$(x_t,a_t,x_{t+1})$}
\psfrag{ftilde}{\tiny\hspace{-0.2cm}$\tilde f(x,a),c(x,a)$}
\includegraphics[width=0.5\textwidth]{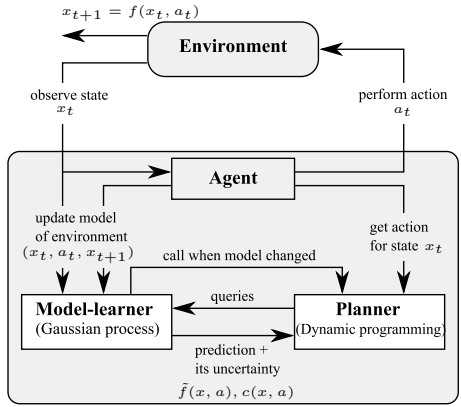}
\end{center}
\caption{High-level overview of the GP-RMAX framework}
\label{fig:2}
\end{figure}
In the last section we have seen how, for a continuous state space, optimal behavior of an agent can be obtained in 
a numerically robust way, given that the transition function $x_{t+1}=f(x_t,a_t)$ is known.\footnote{Remember our 
working assumption: reward as a performance criterion is externally given and does not need to be estimated
by the agent. Also note that discretization (even with more advanced methods like adaptive or sparse grids)
is likely to be feasible only in state spaces with low to medium dimensionality. Breaking the curse of dimensionality
is an open research problem.}

For model-based RL we are now interested in solving the same problem for the case that the transition function is not
known. Instead, the agent has to {\em interact} with the environment, and only use the samples it observes to compute
optimal behavior. Our goal in this paper is to develop a learning framework where this number is kept as small as
possible. This will be done by using the samples to learn an estimate $\tilde f(x,a)$ of $f(x,a)$ and then use this
estimate $\tilde f$ in place of $f$ in the numerical procedure outlined in the previous section.

\subsection{Overview}
%
A sketch of our architecture is shown in Figure~\ref{fig:2}. GP-RMAX consists of 
the two parts model learning and planning which are interwoven for online learning. 
The model-learner estimates the dynamics of the environment from the sample transitions 
the agent experiences while interacting with the environment. The planner is used to find 
the best possible action, given the current model. As the predictions of the model-learner 
become increasingly more accurate, the actions derived from the planner become increasingly 
closer to optimal. Below is a high-level overview of the algorithm:

\begin{itemize}
\item {\bf Input:} 
	\begin{itemize}
	\item Reward function $r(x,a)$
	\item Discount factor $\gamma$
	\item Performance parameters:
		\begin{itemize}
		\item planning and model-update frequency $K$
		\item model accuracy $\delta^M_1,\delta^M_2$ (stopping criterion for model-learning)
		\item discretization of planner $N$
		\end{itemize}  
	\end{itemize} \medskip

\item{\bf Initialize:}
	\begin{itemize}
	\item Model $\mathcal M_1$, Q-function $\mathcal Q_1$, observed transitions $\mathcal D_1$
	\end{itemize}\medskip
	
\item{\bf Loop:} $t=1,2,\ldots$
	\begin{itemize}
	\item {\bf Interact with system:}
		\begin{itemize}
		\item observe current state $x_t$
		\item choose action $a_t$ greedy with respect to $\mathcal Q_t$
			\[ a_t=\argmax_{a'} \mathcal Q_t(x_t,a') \]
		\item execute action $a_t$, observe next state $x_{t+1}$, store transition $\mathcal D_{t+1}=\mathcal D_t \cup \{x_t,a_t,x_{t+1} \}$
		\end{itemize}
	\end{itemize}\medskip
	
	\begin{itemize}
	\item {\bf Model learning:} (see Section~\ref{sec:Model learning with GPs})
		\begin{itemize}
		\item only every $K$ steps, and only if $\mathcal M_t$ is not sufficiently exact (as determined by evaluating the stopping criterion)
			\begin{itemize}
			\item $\mathcal M_{t+1}=\mathrm{update\_model}\, (\mathcal M_t,\mathcal D_{t+1})$
			\item $\mathrm{evaluate\_stopping\_criterion}\, (\mathcal M_{t+1},\mathcal M_t,\delta^M_1,\delta^M_2)$
			\end{itemize}
		\item else 
			\begin{itemize}
			\item $\mathcal M_{t+1}=\mathcal M_t$
			\end{itemize}
		\end{itemize}
	\end{itemize}\medskip
	
	\begin{itemize}
	\item {\bf Planning with model:} (see Section~\ref{sec:Planning with a model})
		\begin{itemize}
		\item only every $K$ steps, and only if $\mathcal M_t$ is not sufficiently exact (as determined by evaluating the stopping criterion)
			\begin{itemize}
			\item $\mathcal Q_{t+1}=\mathrm{augmented\_value\_iteration}\, (\mathcal{Q_t},\mathcal M_{t+1},@r(x,u),\gamma,N)$
			\end{itemize}
		\item else 
			\begin{itemize}
			\item $\mathcal Q_{t+1}=\mathcal Q_t$
			\end{itemize}
		\end{itemize}
	\end{itemize}		
\end{itemize}

Next, we will explain in more detail how each of the two functional modules ``model-learner'' and  
``planner'' is implemented.

\subsection{Model learning with GPs}
\label{sec:Model learning with GPs}
In essence, estimating $\tilde f$ from samples is a regression problem. While in theory any 
nonlinear regression algorithm could serve this purpose, we believe that GPs are particularly well-suited: 
(1) being non-parametric means great modeling flexibility; (2) setting the 
hyperparameters can be done automatically (and in a principled way) via optimization of the marginal likelihood
and allows automatic determination of relevant  inputs; and (3) GPs provide a natural way to determine the 
uncertainty of its predictions which will be used to guide exploration. Furthermore, uncertainty in GPs is {\em supervised}
in that it depends on the target function that is estimated (because of (2)); other methods only consider the density
of the data ({\em unsupervised}) and will tend to overexplore if the target function is simple.
 
Assume we have observed a number of transitions, given as triplets of state, performed action, and resulting successor
state, e.g., $\mathcal D=\{x_t,a_t,x_{t+1}\}_{t=1,2,\ldots}$ where $x_{t+1}=f(x_t,a_t)$. Note that $f$ is a
$d$-dimensional function, $f(x_t,a_t)=\bigl[f_1(x_t,a_t),\ldots,f_d(x_t,a_t)\bigr]^T$. Instead of trying to estimate
$f$ directly (which corresponds to absolute transitions), we try to estimate the relative change $x_{t+1}-x_t$ as in 
\cite{jong07}. 
The effect of each action on each state variable will be treated independently: we train multiple univariate GPs and 
combine the individual predictions afterwards. Each individual $\mathcal{GP}_{ij}$ is trained in the respective 
subset of data in $\mathcal D$, e.g., $\mathcal{GP}_{ij}$ is trained on all $x_t$ as input, and $x_{t+1}^{(i)}-x_t^{(i)}$ 
as output, where $a_t=j$. Each individual $\mathcal{GP}_{ij}$ has its own set of hyperparameters obtained from optimizing 
the marginal likelihood.

The details of working\footnote{There is also the problem of implementing GPs {\em efficiently} when dealing with a 
possible large number of data points. For the lack of space we can only sketch our particular implementation, 
see \cite{jqc07approxpg} for more detailed information. Our GP implementation is based on the {\em subset of regressors}
approximation. The elements of the subset are chosen by a stepwise greedy procedure aimed at minimizing
the error incurred from using a low rank approximation (incomplete Cholesky decomposition). Optimization of the
likelihood is done on random subsets of the data of fixed size. To avoid a degenerate predictive variance,
the {\em projected process} approximation was used.} with GPs can be found in \cite{raswil06gp}; using GPs to learn 
a model for RL was previously also studied in \cite{deis09gpdp} (for offline RL and without uncertainty-guided exploration). 
One characteristic of GPs is that their functional form is given in terms of a 
parameterized covariance function. Here we use the squared exponential,
\begin{equation*}
k(x,x';v_0,b,\vec\theta)=v_0\exp \Bigl\{-0.5(x-x')^T \Omega (x-x') \Bigr\} +b,
\end{equation*}
where matrix $\Omega$ is either one of the following: (1) $\Omega=\theta I$ (uniform), 
(2) $\Omega=\diag(\theta_1,\ldots,\theta_d)$ (axis aligned ARD), (3) $\Omega=M_k M_k^T$
(factor analysis). Scalars $v_0,b$ and the ($\Omega$-dependent number of) entries of $\vec\theta$
constitute the hyperparameters of the GP and are adapted from the training data 
(likelihood optimization). Note that variant (2) and (3) implement {\em automatic
relevance determination}: relevant inputs or linear projections of inputs are automatically 
identified, whereby model complexity is reduced and generalization sped up.

%

Once trained, for any testpoint $x$, $\mathcal{GP}_{ij}$ provides a distribution over target values, 
$\mathcal N(\mu_{ij}(x),\sigma_{ij}^2(x))$, with mean $\mu_{ij}(x)$ and variance $\sigma_{ij}^2(x)$ (exact formulas for $\mu$ and $\sigma$
can be found in \cite{raswil06gp}). Each individual mean $\mu_{ij}$ predicts the change in the $i$-th coordinate 
of the state under the $j$-th action. Each individual variance $\sigma_{ij}^2$ can be interpreted as the associated 
uncertainty; it will be close to $0$ if $\mathcal{GP}_{ij}$ is certain, and close to $k(x,x)$ if it is 
uncertain (the value of $k(x,x)$ depends on the hyperparameters of $\mathcal{GP}_{ij}$). Stacking the individual 
predictions together, our model-learner produces in summary
\begin{equation}
\tilde f(x,a):=\begin{bmatrix} x^{(1)} \\ \vdots \\ x^{(d)} \end{bmatrix} 
+ \begin{bmatrix} \mu_{1a}(x) \\ \vdots \\ \mu_{da}(x) \end{bmatrix},
\qquad
c(x,a):=\max_{i=1,\ldots, d} \, \big(\mathrm{normalize}_{ia} (\sigma^2_{ia}) \bigr),
\label{eq:tildef}
\end{equation} 
where $\tilde f(x,a)$ is the predicted successor state and $c(x,a)$ the associated uncertainty (taken as maximum over the 
normalized per-coordinate uncertainties, where normalization ensures that the values lie between $0$ and $1$).

\subsection{Planning with a model}
\label{sec:Planning with a model}
At any time $t$, the planner receives as input model $\mathcal M_t$. For any state $x$ and action $a$,
model $\mathcal M_t$ can be evaluated to ``produce'' the transition $\tilde f(x,a)$ along with normalized
scalar uncertainty $c(x,a) \in [0,1]$, where $0$ means maximally certain and $1$ maximally uncertain
(see Section~\ref{sec:Model learning with GPs})  

Let $\Gh$ be the discretization of the state space $\mathcal X$ with nodes $\xi_i$, $i=1,\ldots,N$. We 
now solve the planning stage by plugging $\tilde f$ into the procedure described in 
Section~\ref{sec:numerical solution}. First, we compute $\tilde z_i^a=\tilde f(\xi_i,a)$, 
$c(\xi_i,a)$ from \eqref{eq:tildef} and the associated interpolation coefficients $w_{ij}^a$
from \eqref{eq:3} for each node $\xi_i$ and action $a$. 

Let $C^a$ denote the $N \times 1$ vector corresponding to the uncertainties, $[C^a]_i=c(\xi_i,a)$; and
$R^a$ be the $N\times 1$ vector corresponding to the rewards, $[R^a]_i=r(\xi_i,a)$. To solve the discretized
Bellman equation in Eq.~\eqref{eq:4}, we perform basic Jacobi iteration:
\begin{itemize}
\item {\bf Initialize} $[Q_0^a]_i$, $i=1,\ldots,N$, $a=1,\ldots,|A|$
\item {\bf Repeat} for $k=0,1,2,\ldots$ 
\begin{equation}
[Q_{k+1}^a]_i = [R^a]_i + \gamma \max_{a'} \left\{ \sum_{j=1}^N w_{ij}^a [Q_k^{a'}]_j \right\} \quad \forall i,a
\label{eq:7}
\end{equation} 
{\bf until} $|Q_{k+1}^a-Q_k^a|_{\infty}<tol$, $\forall a$, or a maximum number of iterations is reached.
\end{itemize}

To reduce the number of iterations necessary, we adapt Gr\"une's {\em increasing coordinate algorithm} \cite{Gruene97}
to the case of Q-functions: instead of Eq.~\eqref{eq:7}, we perform updates of the form
\begin{equation}
\tag{\ref{eq:7}'}
[Q_{k+1}^a]_i = [1-\gamma w_{ii}^a]^{-1} \left( 
[R^a]_i + \gamma \max_{a'} \left\{ \sum_{j=1,j\ne i}^N w_{ij}^a [Q_k^{a'}]_j \right\}
\right).
\label{eq:7'}
\end{equation} 
In \cite{Gruene97} it was proved that Eq.~\eqref{eq:7'} converges to the same fixed point as Eq.~\eqref{eq:7}, and
it was empirically demonstrated that convergence can occur in significantly fewer iterations.
The exact reduction is problem-dependent, savings will be greater for small $\gamma$ and large cells 
where self-transitions occur (i.e., $\xi_i$ is among the vertices of the cell enclosing $\tilde z_i^a$).

To implement the ``optimism in the face of uncertainty'' principle, that is, to make the agent explore regions
of the state space where the model predictions are uncertain, we employ the heuristic modification of the
Bellman operator which was suggested in \cite{nouri08multiresolution} and shown to perform well. Instead of Eq.~\eqref{eq:7'},
the update rule becomes  
\begin{multline}
\tag{\ref{eq:7}''}
[Q_{k+1}^a]_i = (1-[C^a]_i) [1-\gamma w_{ii}^a]^{-1} \left( 
[R^a]_i + \gamma \max_{a'} \left\{ \sum_{j=1,j\ne i}^N w_{ij}^a [Q_k^{a'}]_j \right\}
\right)+\\
+[C^a]_i V_\text{MAX}
\label{eq:7''}
\end{multline} 
where $V_\text{MAX}:=R_\text{MAX}/(1-\gamma)$. Eq.~\eqref{eq:7''} can be seen as a generalization of the
binary uncertainty in the original RMAX paper to continuous uncertainty; whereas in RMAX a state was either
``known'' (sufficiently explored), in which case the unmodified update was used, or ``unknown'' (not sufficiently
explored), in which case the value $V_\text{MAX}$ was assigned, here the shift from exploration to exploitation
is more gradual.

Finally we can take advantage of the fact that the planning function will be called many times during the
process of learning. Since the discretization $\Gh$ is kept fixed, we can reuse the final Q-values obtained
in one call to plan as initial values for the next call to plan. Since updates to the model often affect
only states in some local neighborhood (in particular in later stages), the number of necessary iterations
in each call to planning will be further reduced.
 
A summary of our model-based planning function is shown below.

\begin{itemize}
\item {\bf Input:} 
	\begin{itemize}
	\item Model $\mathcal M_t$, initial $[Q_0^a]_i$, $i=1,\ldots,N$, $a=1,\ldots,|A|$
	\end{itemize} \medskip
\item{\bf Static inputs:}
	\begin{itemize}
	\item Grid $\Gh$ with nodes $\xi_1,\ldots,\xi_N$, discount factor $\gamma$, reward function $r(x,a)$ evaluated in nodes giving $[R^a]_i$
	\end{itemize} \medskip
\item{\bf Initialize:}
	\begin{itemize}
	\item Compute $\tilde z_i^a=\tilde f(\xi_i,a)$ and $[C^a]_i$ from $\mathcal M_t$ (see Eq.~\eqref{eq:tildef})
	\item Compute weights $w_{ij}^a$ for each $\tilde z_i^a$ (see Eq.~\eqref{eq:3})
	\end{itemize}\medskip
\item{\bf Loop:}
	\begin{itemize}
	\item Repeat update Eq.~\eqref{eq:7''} until $|Q_{k+1}^a-Q_k^a|_{\infty}<tol$, $\forall a$, or the maximum number
	of iterations is reached.
	\end{itemize}
\end{itemize}

\section{Experiments}
We now examine the online learning performance of GP-RMAX in various well-known
RL benchmark domains. 

\subsection{Description of domains}
In particular, we choose the following domains (where a large number of comparative results 
is available in the literature):

\paragraph{\bf Mountain car:} In mountain car, the goal is to drive an underpowered car from the bottom 
of a valley to the top of one hill. The car is not powerful enough to climb the hill directly, instead it 
has to build up the necessary momentum by reversing throttle and going up the hill on the opposite side 
first. The problem is $2$-dimensional, state variable $x_1 \in [-1.2,0.5]$ describes the position of 
the car, $x_2 \in [-0.07,0.07]$ its velocity. Possible actions are $a\in \{-1,0,+1\}$. Learning is episodic:
every step gives a reward of $-1$ until the top of the hill at $x_1 \ge 0.5$ is reached. Our experimental
setup (dynamics and domain specific constants) is the same as in \cite{sutton98introduction}, with the 
following exceptions: maximal episode length is $500$ steps, discount factor $\gamma=0.99$ and every episode 
starts with the agent being at the bottom of the valley with zero velocity, $x_\text{start}=(-\pi/6,0)$.

\paragraph{\bf Inverted pendulum:} The next task is to swing up and stabilize a single-link inverted 
pendulum. As in mountain car, the motor does not provide enough torque to push the pendulum up in 
a single rotation. Instead, the pendulum needs to be swung back and forth to gather energy, before being
pushed up and balanced. This creates a more difficult, nonlinear control problem. The state space is 
$2$-dimensional, $\theta \in [-\pi,\pi]$ being the angle, $\dot \theta \in [-10,10]$ the angular velocity. Control
force is discretized to $a\in \{-5,-2.5,0,+2.5,+5\}$ and held constant for $0.2$sec. Reward is defined as
$r(x,a):=-0.1x_1^2 -0.01x_2^2 -0.01a^2$. The remaining experimental setup (equations of motion and domain 
specific constants) is the same as in \cite{deis09gpdp}. The task is made episodic by resetting the system 
every $500$ steps to the initial state $x_\text{start}=(0,0)$. Discount factor $\gamma=0.99$.

\paragraph{\bf Bicycle:} Next we consider the problem of balancing a bicycle that rides at a constant 
speed \cite{Ernst05},\cite{Lagoudakis03}.
The problem is $4$-dimensional: state variables are the roll angle $\omega \in [-12\pi/180,12\pi/180]$, roll rate
$\dot \omega \in [-2\pi,2\pi]$, angle of the handle bar $\alpha \in [-80\pi/180,80\pi/180]$, and the angular 
velocity $\dot\alpha \in [-2\pi, 2\pi]$. The action space is inherently $2$-dimensional (displacement of rider
from the vertical and turning the handlebar); in RL it is usually discretized into $5$ actions. Our experimental 
setup so far is similar to \cite{Ernst05}. To allow a more conclusive comparison of performance, instead of just being
able to keep the bicycle from falling, we define a more discriminating reward $r(x,a)=-x_1^2$, and $r(x,a)=-10$
for $|x_1|<12\pi/180$ (bicycle has fallen). Learning is episodic: every episode starts in 
one of two (symmetric) states close to the boundary from where recovery is impossible: 
$x_\text{start}=(10\pi/180,0,0,0)$ or $x_\text{start}=(-10\pi/180,0,0,0)$, and proceeds for $500$ steps or until
the bicycle has fallen. Discount factor $\gamma=0.98$.

\paragraph{\bf Acrobot:} Our final problem is the acrobot swing-up task \cite{sutton98introduction}. The goal is to swing up the
tip of the lower link of an underactuated two-link robot over a given height (length of first link). Since only the lower link 
is actuated, this is a rather challenging problem. The state space is $4$-dimensional:
$\theta_1 \in [-\pi,\pi]$, $\dot \theta_1 \in [-4\pi,4\pi]$, $\theta_2 \in [-\pi,\pi]$, $\dot \theta_2 \in [-9\pi,9\pi]$.
Possible actions are $a \in \{-1,+1\}$. Our experimental setup and implementation of state transition dynamics is similar
to \cite{sutton98introduction}. The objective of learning is to reach a goal state as quickly as possible, thus $r(x,a)=-1$ for 
every step. The initial state for every episode is $x_\text{start}=(0,0,0,0)$. An episode ends if either a goal state is 
reached or $500$ steps have passed. The discount factor was set to $\gamma=1$, as in \cite{sutton98introduction}. 

\subsection{Results}
%
We now apply our algorithm GP-RMAX to each of the four problems. The granularity 
of the discretization $\Gh$ in the planner is chosen such that for the $2$-dimensional problems,
the loss in performance due to discretization is negligible. For the $4$-dimensional problems, we
ran offline trials with the true transition function to find the best compromise of granularity 
and computational efficiency. As result, we use a $100 \times 100$ grid for mountain car and 
inverted pendulum, a $20 \times 20 \times 20 \times 20$ grid for the bicycle balancing task, 
and a $25 \times 25 \times 25 \times 25$ grid for the acrobot. The maximum number of value
iterations was set to $500$, tolerance was $<10^{-2}$. In practice, running the full planning 
step took between 0.1-10 seconds for the small problems, and less than 5 min for the large problems 
(where often more than 50\% of the CPU time was spent on computing the GP predictions in all the nodes of the grid).
Using the planning module offline with the true transition function, we computed the best possible 
performance for each domain in advance. We obtained: mountain car (103 steps), inverted pendulum 
(-18.41 total cost), bicycle balancing (-3.49 total cost), and acrobot (64 steps).\footnote{Note 
that 64 steps is not the optimal solution, \cite{Boone97} demonstrated swing-up with 61 steps.} 

For the GP-based model-learner, we set the maximum size of the subset to $1000$, and ICD tolerance to 
$10^{-2}$. The hyperparameters of the covariance were not manually tuned, but found from the data by 
likelihood optimiziation.
 
Since it would be computationally too expensive to update the model and perform the full planning step
after every single observation, we set the planning frequency $K$ to $50$ steps. 
To gauge if optimal behavior is reached and further learning becomes unnessecary, we monitor the change
in the model predictions and uncertainties between successive updates and stop if both fall below 
a threshold (test points in a fixed coarse grid).
   
We consider the following variations of the base algorithm: (1) {\scshape GP-RMAXexp}, which actively 
explores by adjusting the Bellman updates in Eq.~\eqref{eq:7''} according to the uncertainties produced by 
the GP prediction; (2) {\scshape GP-RMAXgrid}, which does the same but uses binary uncertainty by overlaying 
a uniform grid on top of the state-action space and keeping track which cells are visited; and (3) {\scshape GP-RMAXnoexp}, 
which does not actively explore (see Eq.~\eqref{eq:7'}).  
For comparison, we repeat the experiments using the standard online model-free RL algorithm Sarsa($\lambda$) 
with tile coding \cite{sutton98introduction}, where we consider two different setup of the tilings (one finer and 
one coarser).

\begin{figure}[p]
\begin{center}
\includegraphics[width=0.46\textwidth]{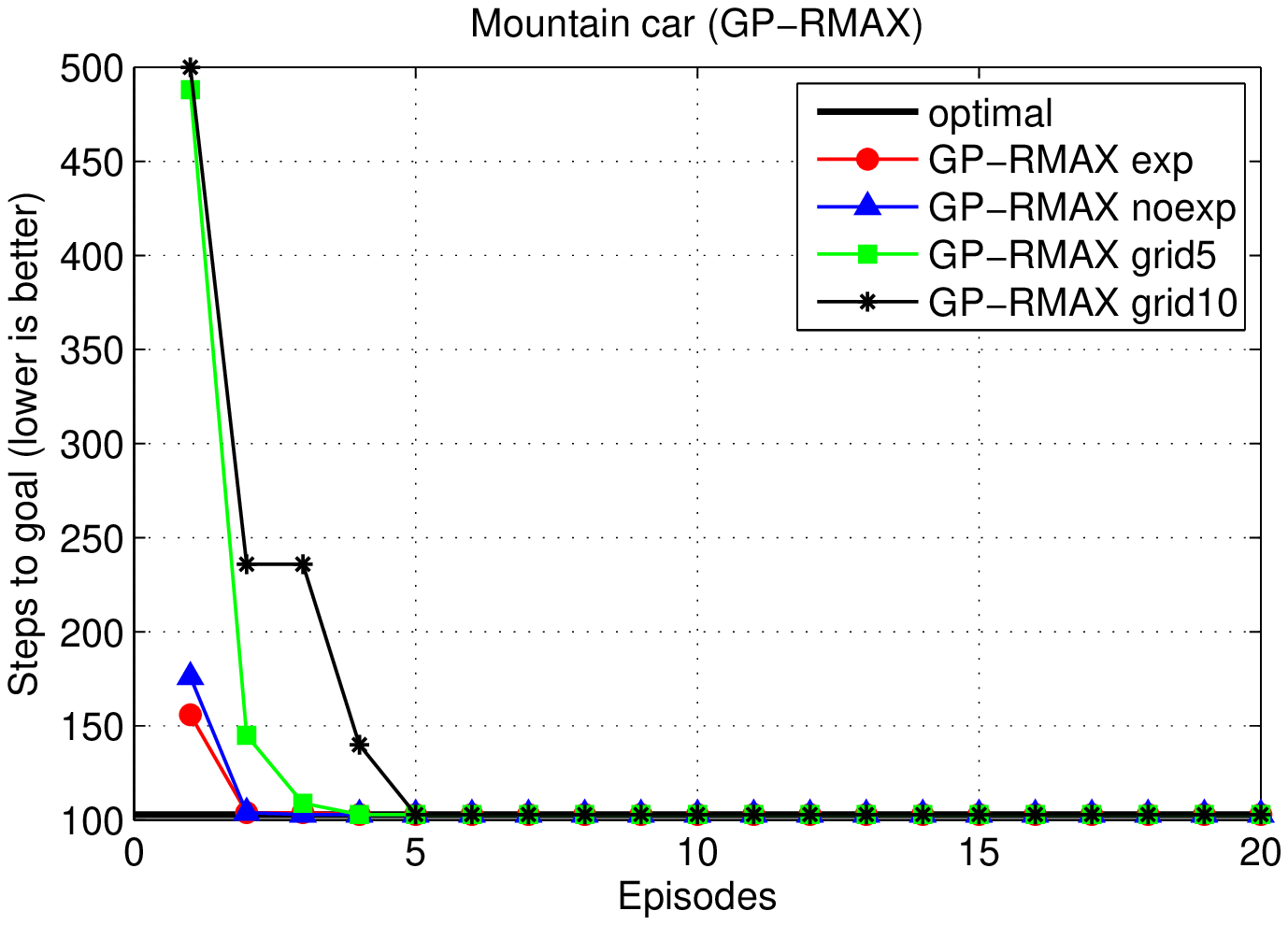}
\includegraphics[width=0.46\textwidth]{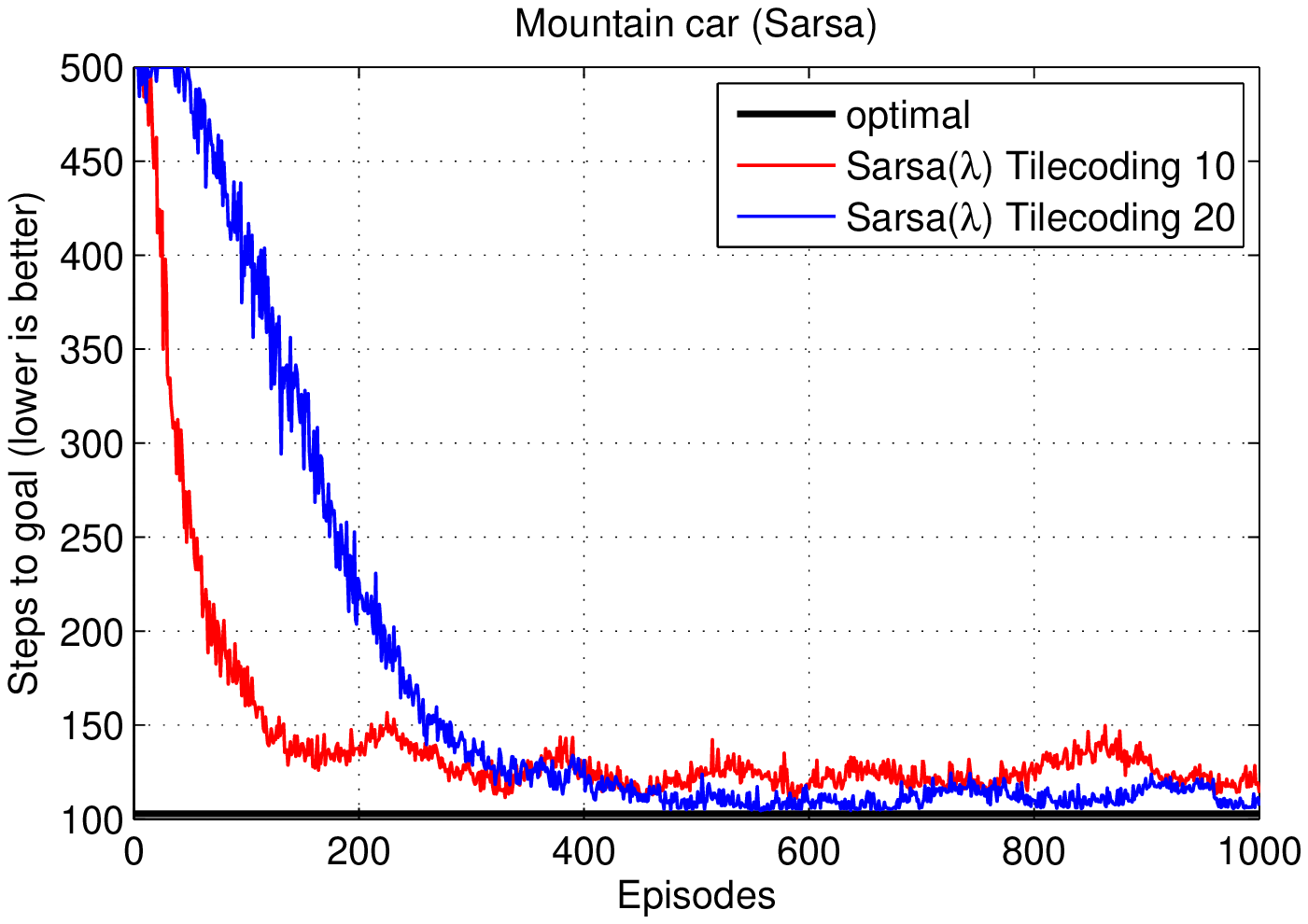}
\end{center}

\begin{center}
\includegraphics[width=0.46\textwidth]{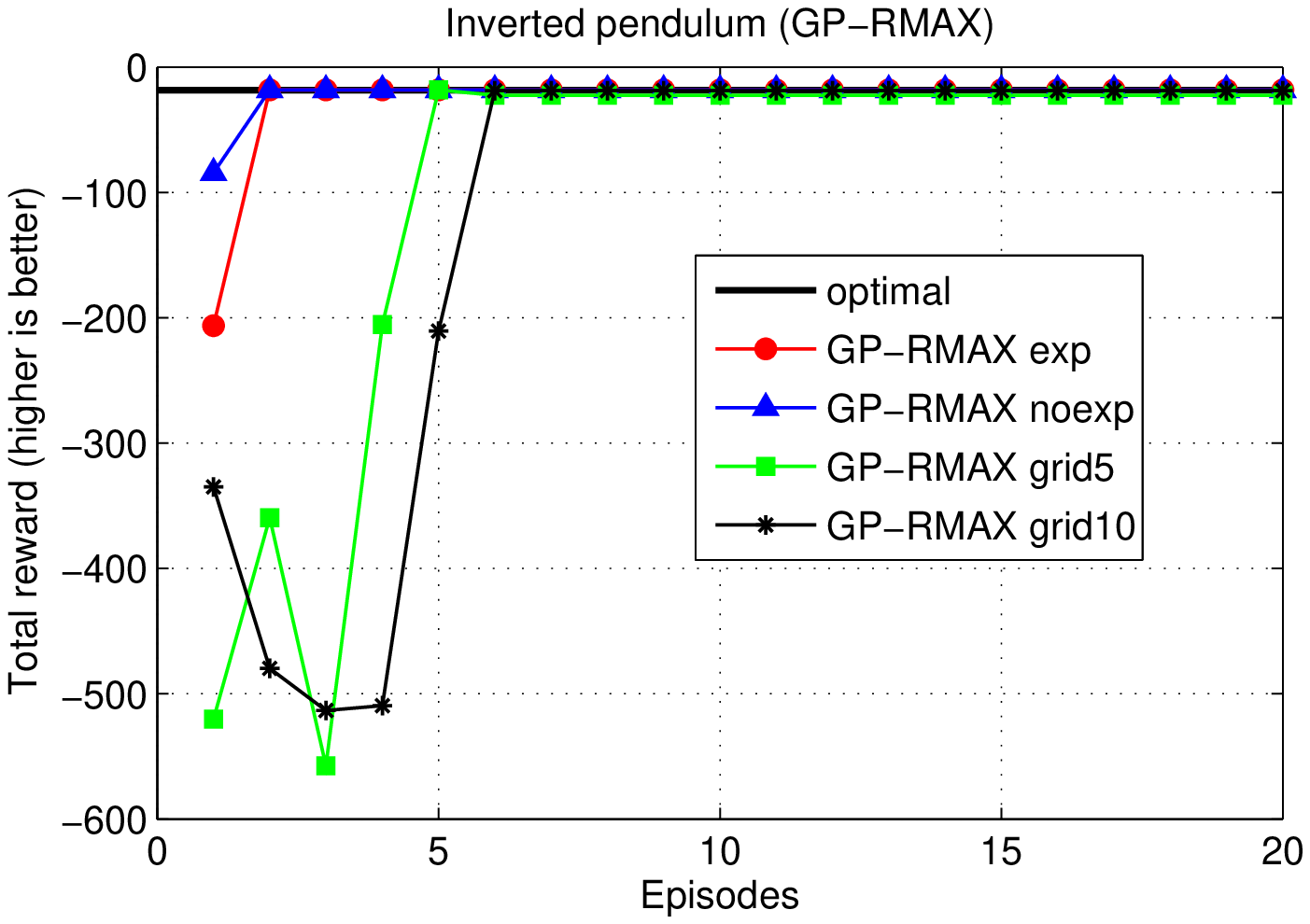}
\includegraphics[width=0.46\textwidth]{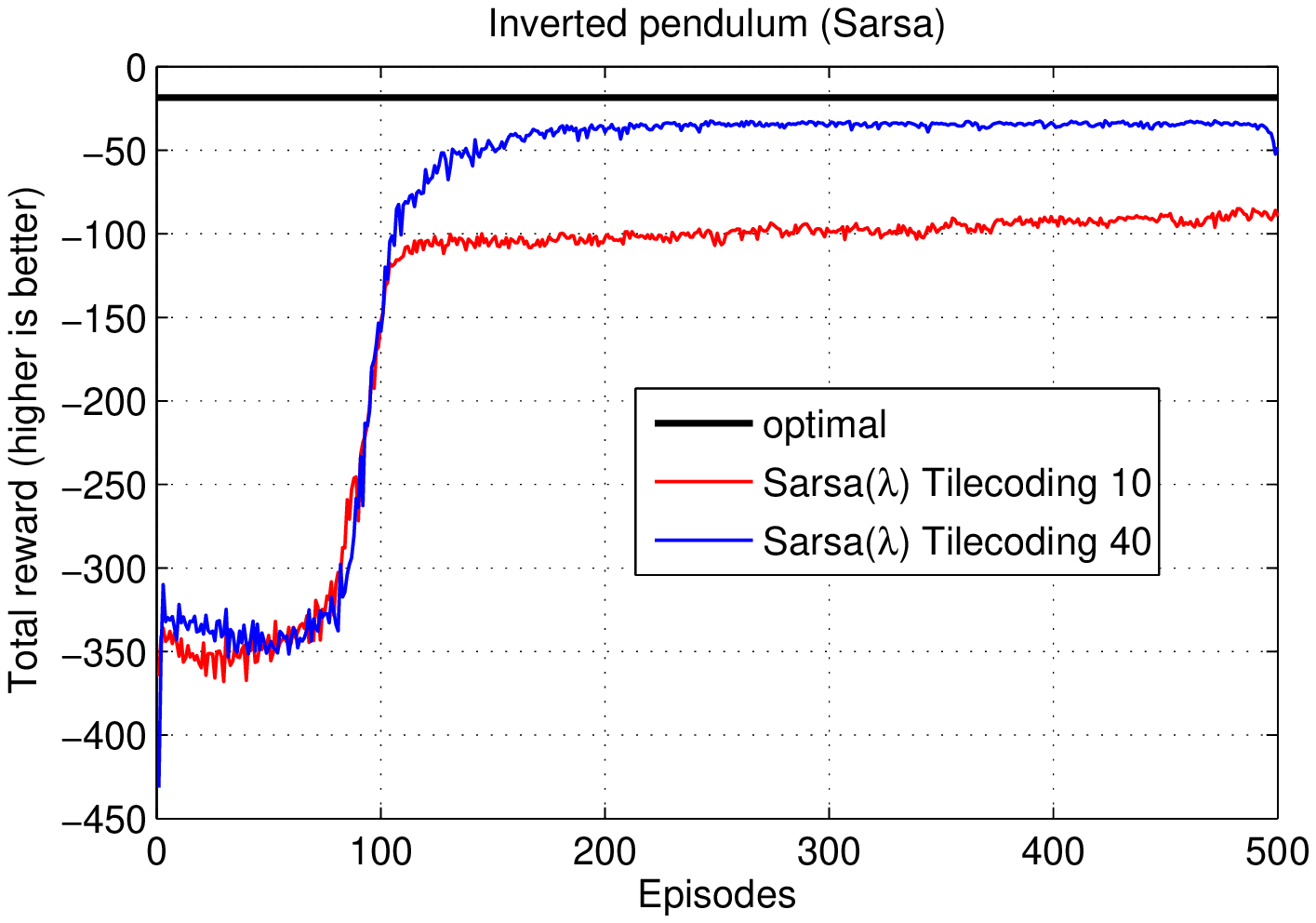}
\end{center}

\begin{center}
\includegraphics[width=0.46\textwidth]{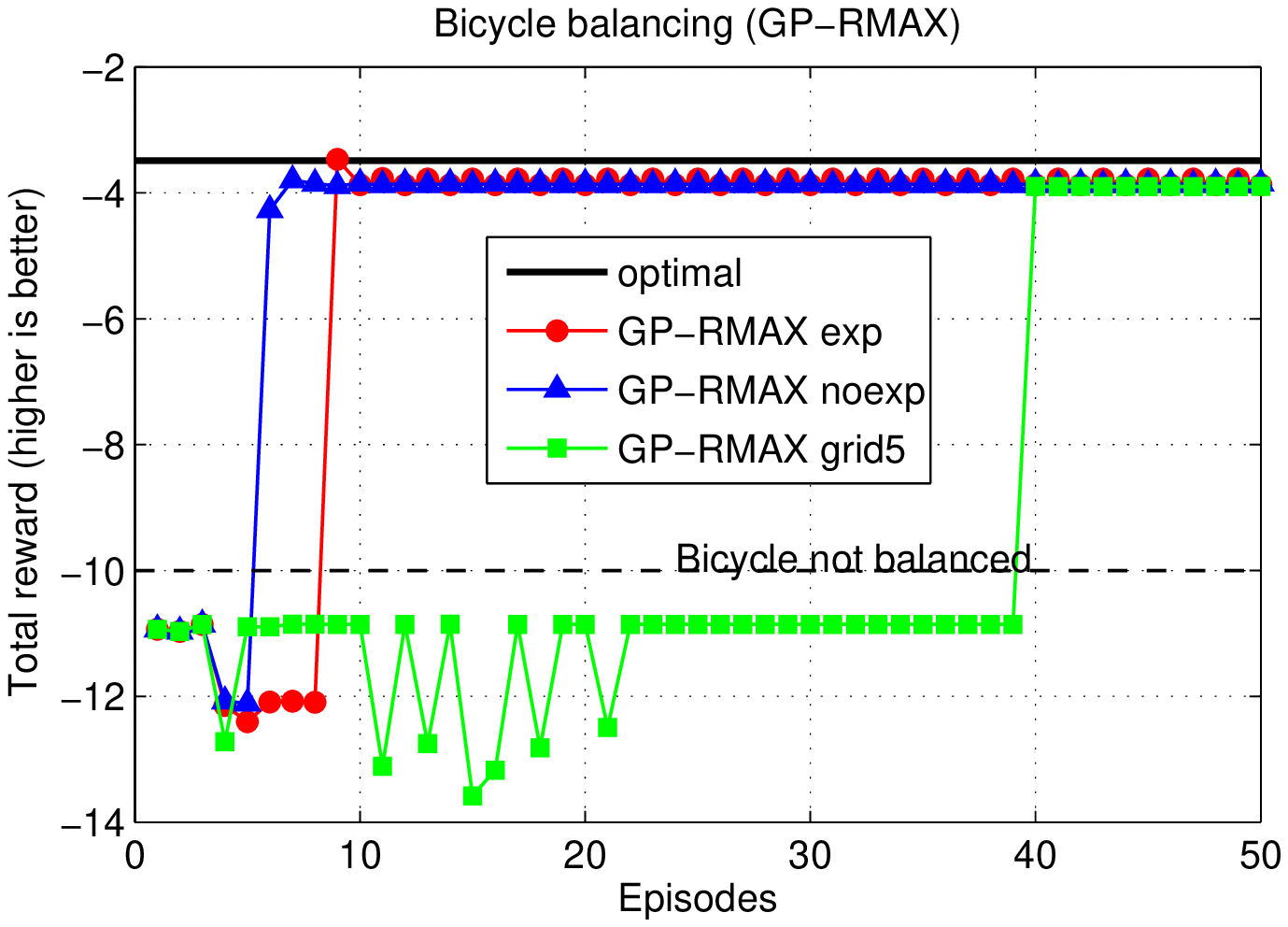}
\includegraphics[width=0.46\textwidth]{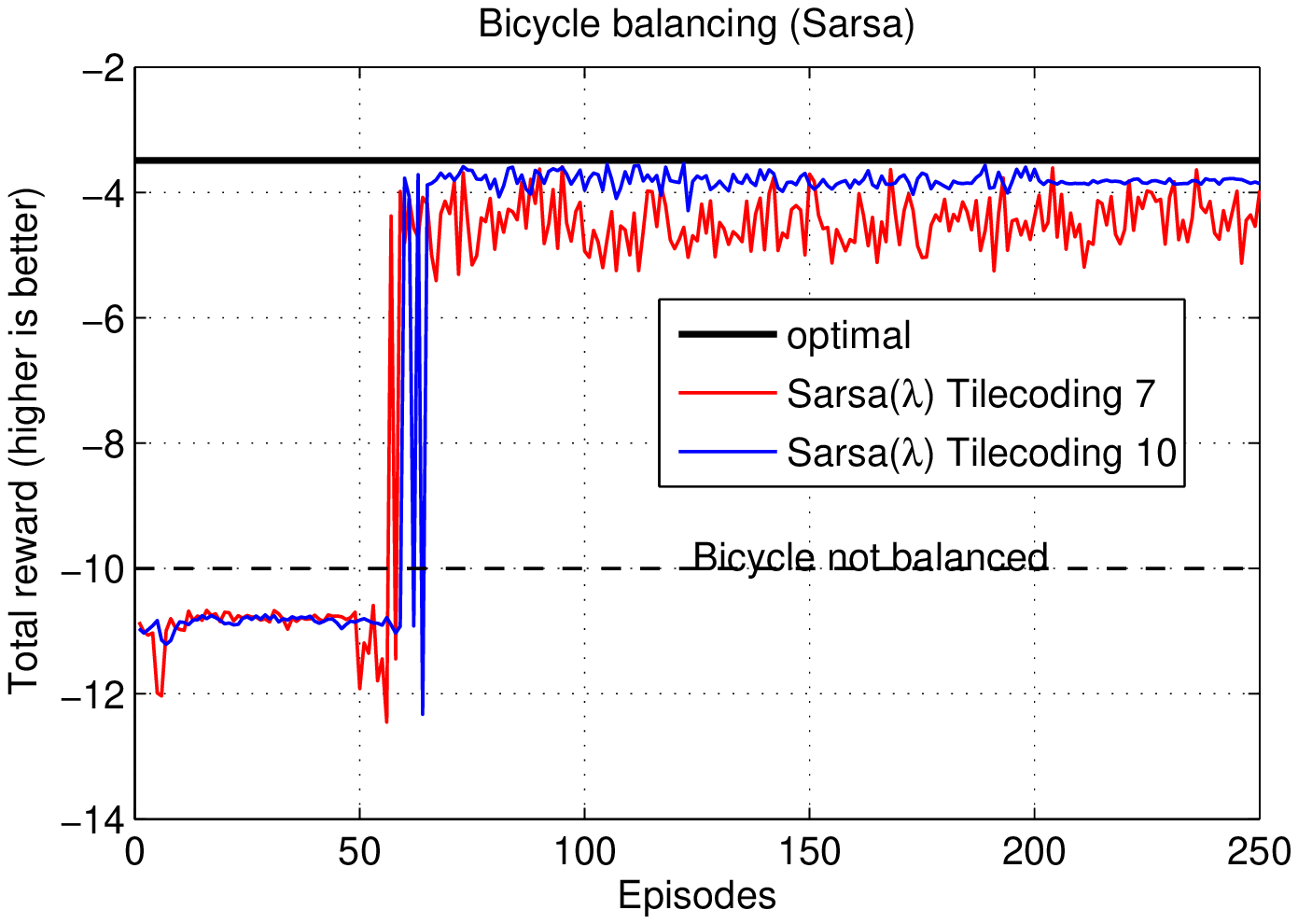}
\end{center}

\begin{center}
\includegraphics[width=0.46\textwidth]{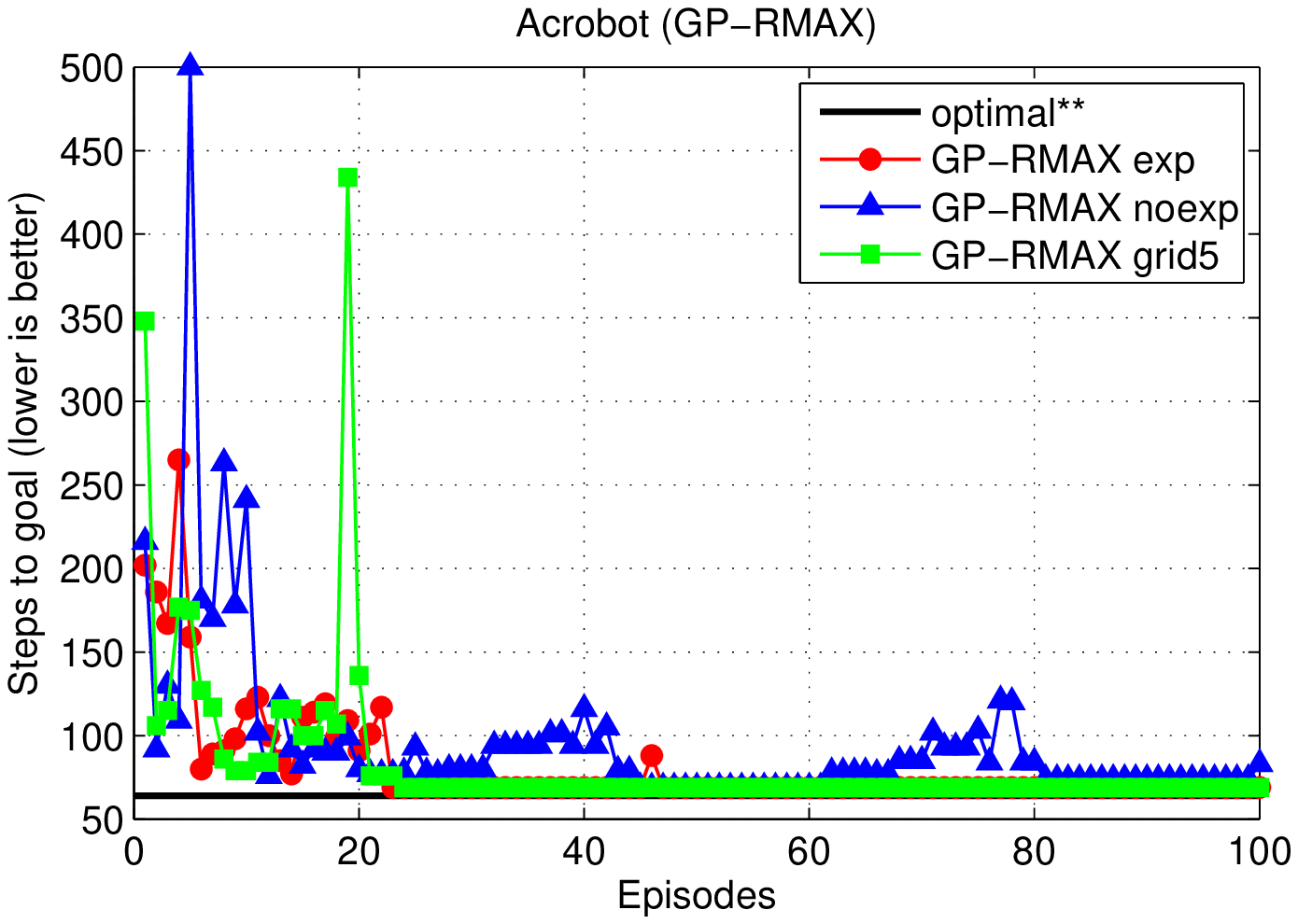}
\includegraphics[width=0.46\textwidth]{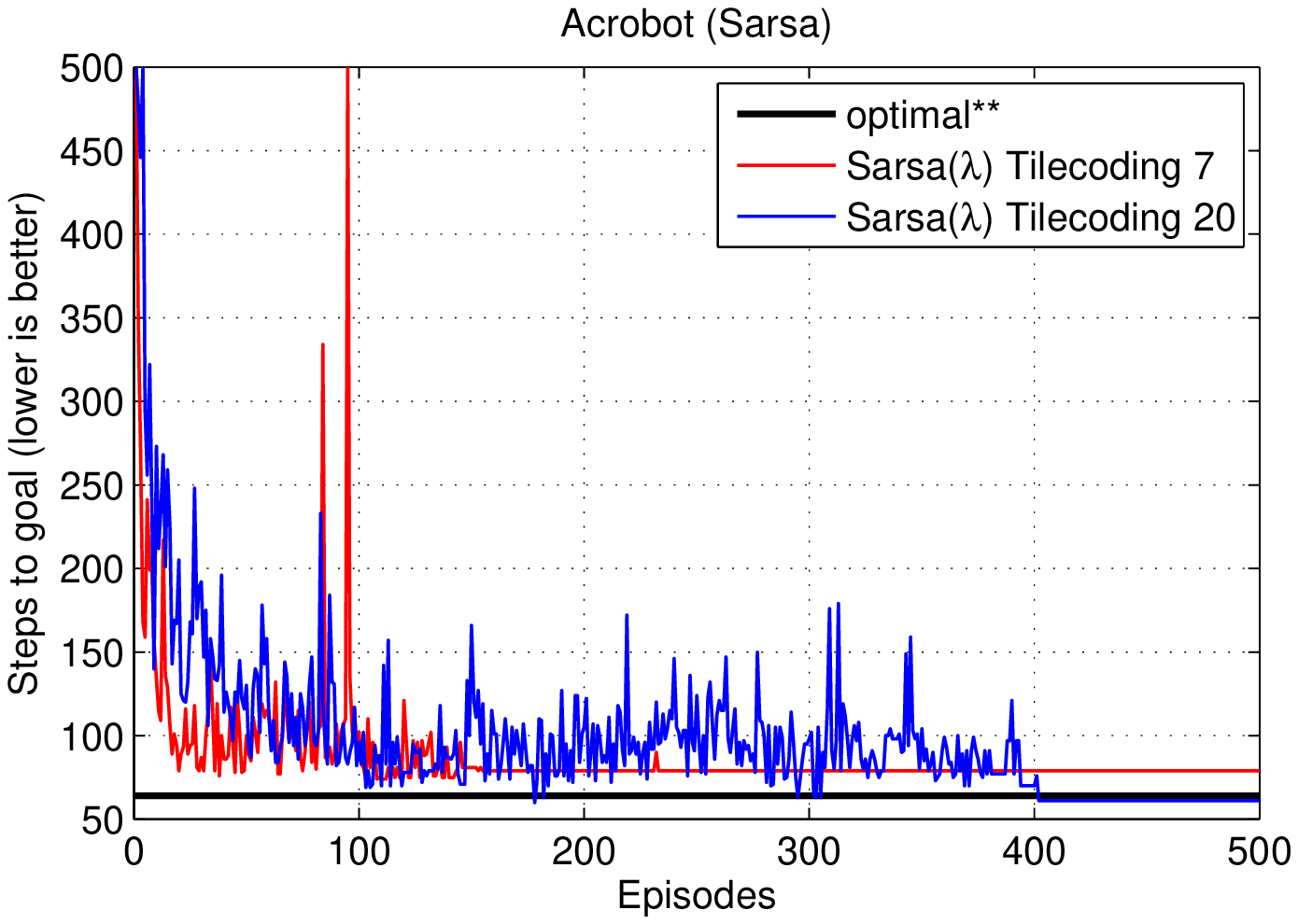}
\end{center}
\caption{Learning curves of our algorithm GP-RMAX (left column) and the standard method Sarsa($\lambda$) 
with tile coding (right column) in the four benchmark domains. Each curve shows the online learning performance 
and plots the total reward as a function of the episode (and thus sample complexity). The black horizontal line 
denotes the best possible performance computed offline. Note the different scale of the x-axis between 
GP-RMAX and Sarsa.} 
\label{fig:4}
\end{figure}

Figure~\ref{fig:4} shows the result of online learning with GP-RMAX and Sarsa. In short, the graphs show us 
two things in particular: (1) GP-RMAX learns very quickly; and (2) GP-RMAX learns
a behavior that is very close to optimal. In comparison, Sarsa($\lambda$) has a much higher sample complexity
and does not always learn the optimal behavior (exception is the acrobot). 
While direct comparison with other high performance RL algorithms, 
such as fitted value iteration \cite{Riedmiller05,Ernst05,nouri08multiresolution}, policy iteration based 
LSPI/LSTD/LSPE \cite{Lagoudakis03,busoniu10onlinelspi,li09LSPIonlineexploration,mein_keepaway2007}, or other kernel-based 
methods \cite{engel2003gptd,deis09gpdp} is difficult, because they are either batch methods or handle exploration
in a more ad-hoc way, from the respective results given in the literature it is clear that for the domains we examined 
GP-RMAX performs relatively well.

\begin{figure}[t!]

\begin{minipage}{0.325\textwidth}
\begin{center}
\includegraphics[width=1.05\textwidth]{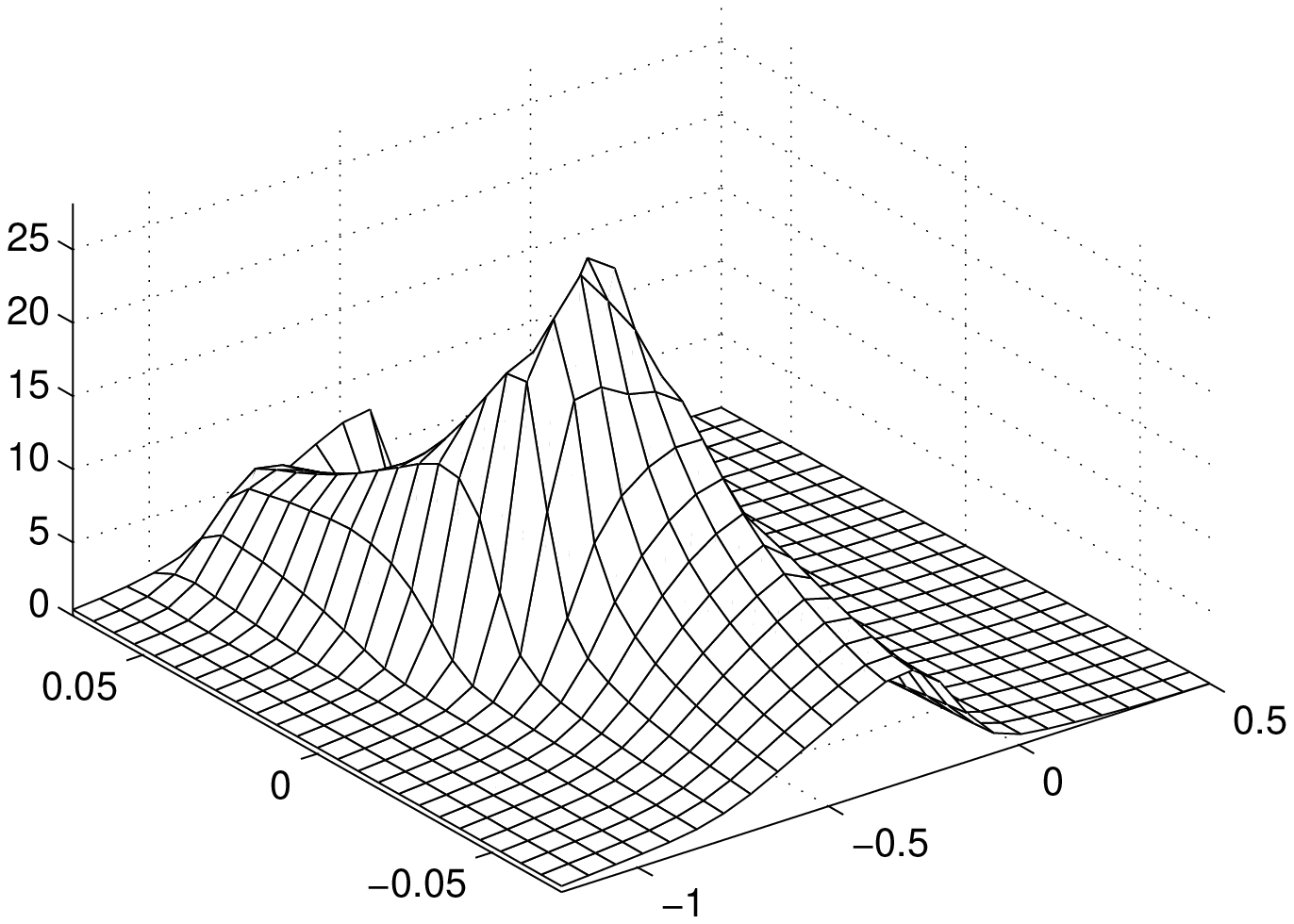}
\includegraphics[width=1\textwidth]{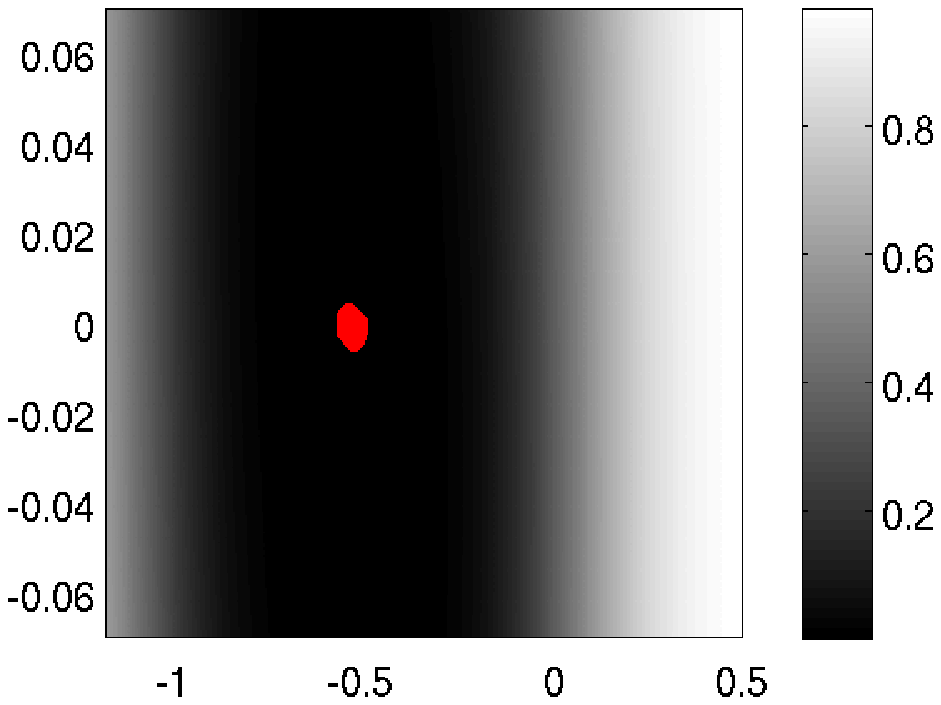}
\centerline{(a) GP after 40 samples}
\end{center}
\end{minipage}
\begin{minipage}{0.325\textwidth}
\begin{center}
\includegraphics[width=1.05\textwidth]{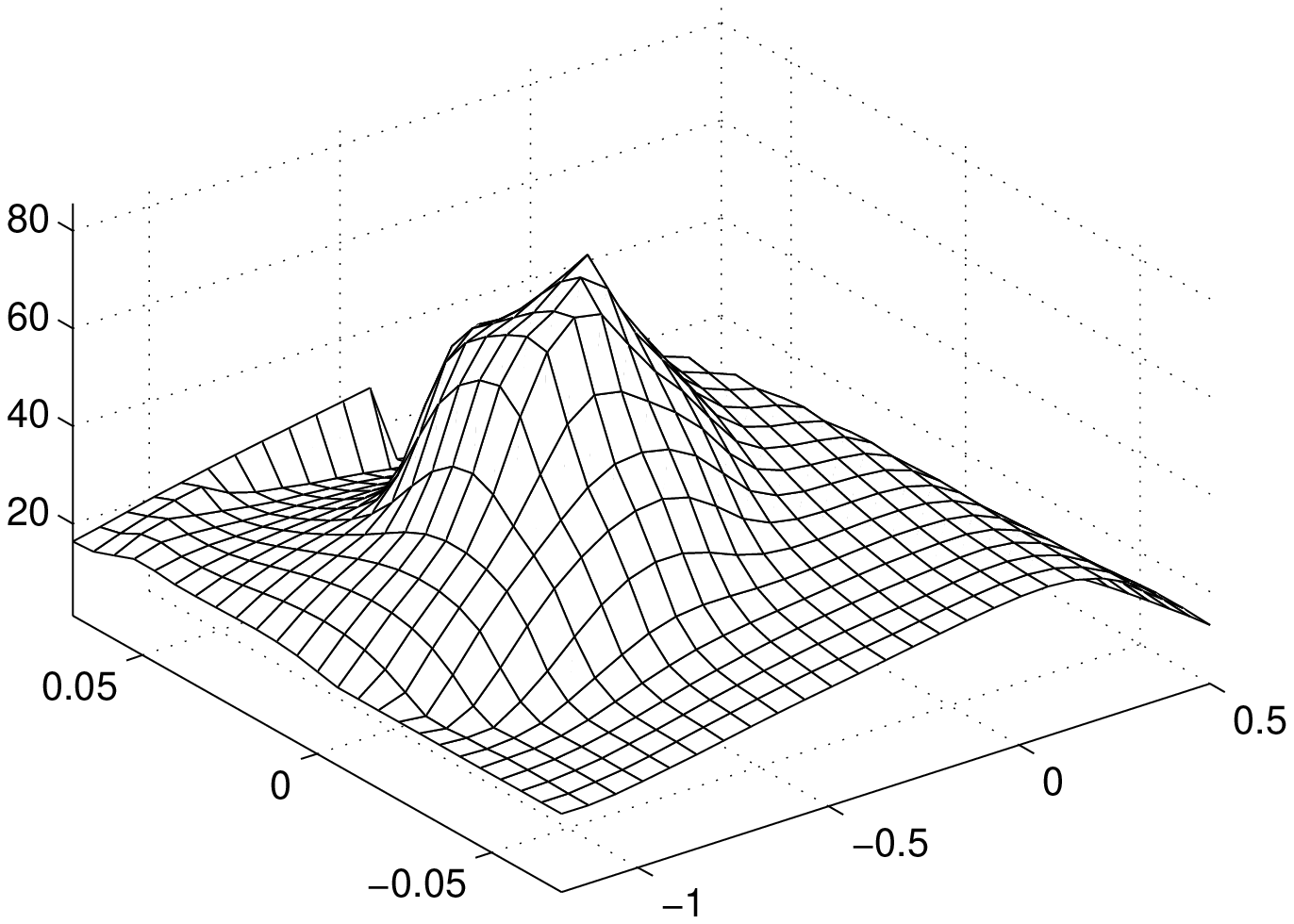}
\includegraphics[width=1\textwidth]{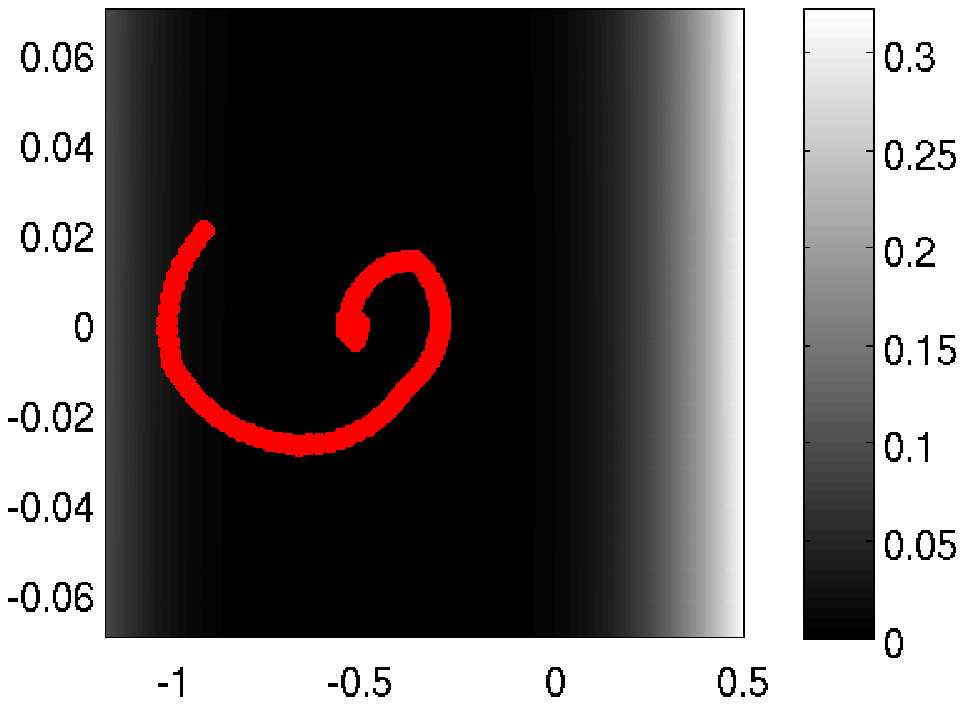}
\centerline{(b) GP after 120 samples}
\end{center}
\end{minipage}
\begin{minipage}{0.325\textwidth}
\begin{center}
\includegraphics[width=1.05\textwidth]{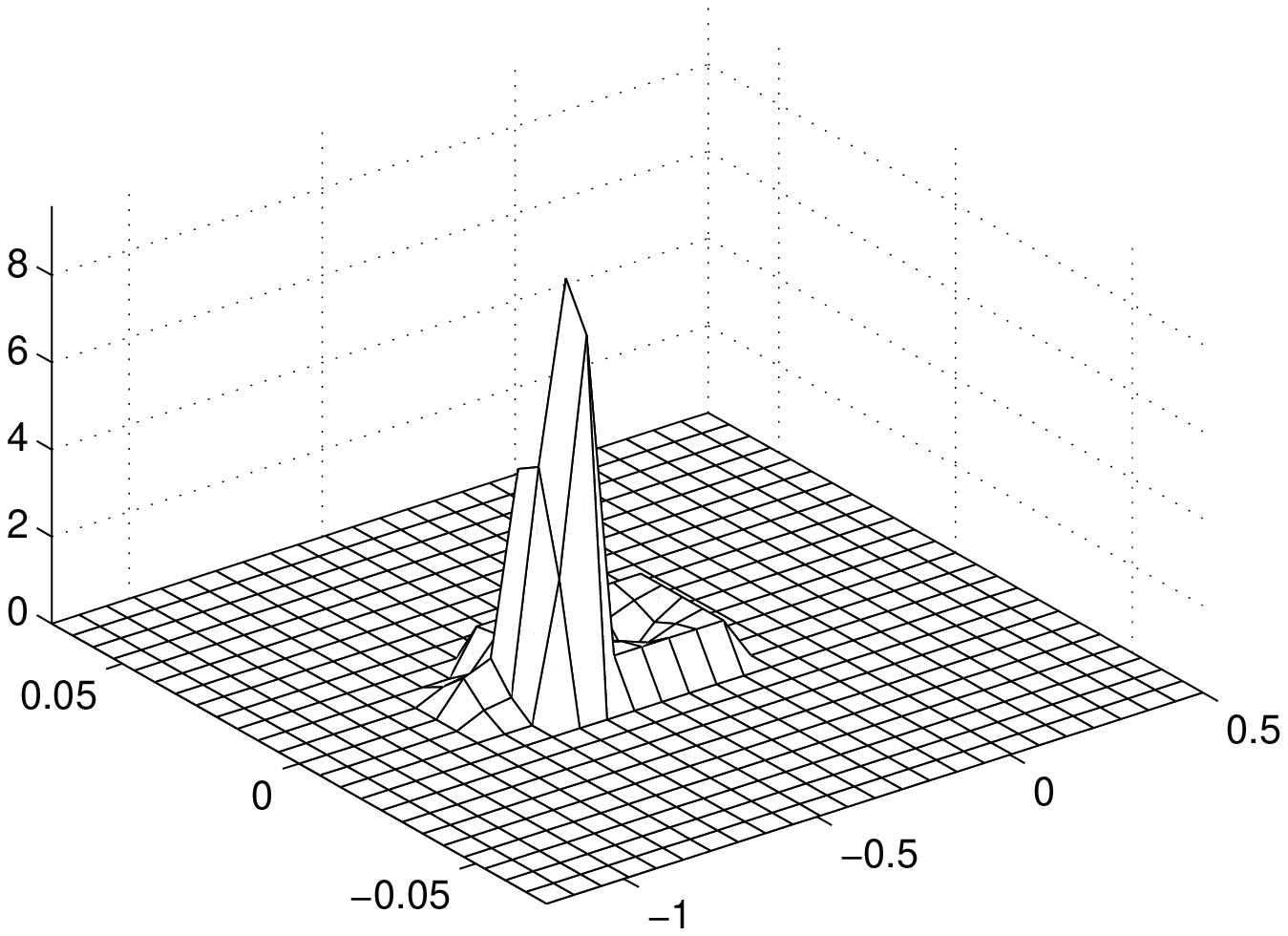}
\includegraphics[width=1\textwidth]{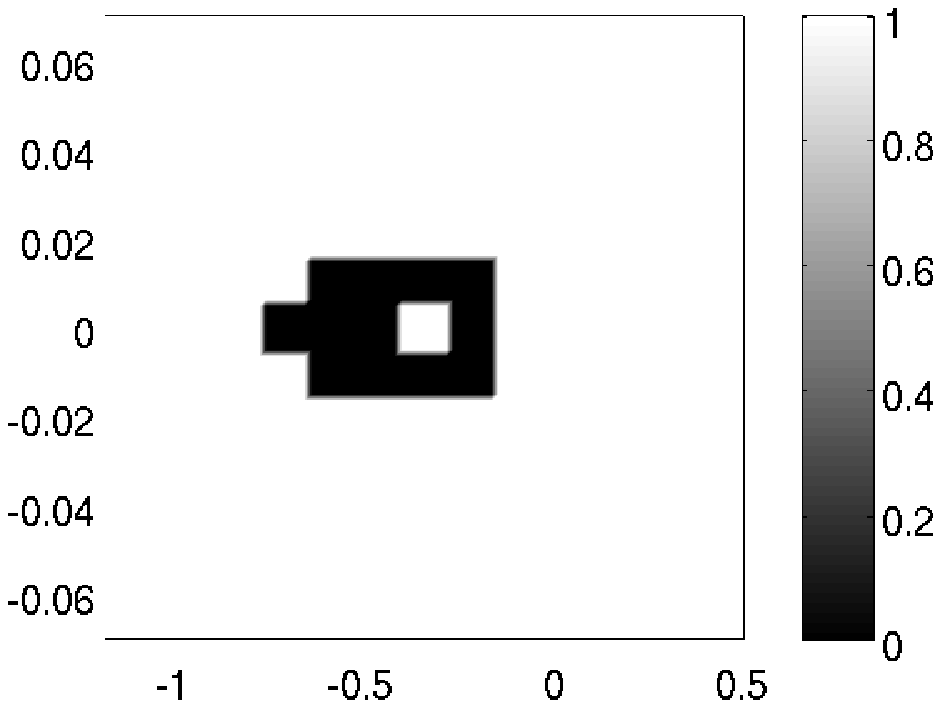}
\centerline{(c) Grid after 120 samples}
\end{center}
\end{minipage}
\caption{Model-learning and propagation of ``knownness'' of state-action pairs with GPs. 
The top row shows the value function that results from applying value iteration with the 
update modified for uncertainty, see Eq.~\eqref{eq:7''}. The bottom row shows the actual 
samples (red circles) and the induced uncertainty of all states: black is 
perfectly ``known'', white is perfectly ``unknown''. Panels (a) and (b) show that with GPs 
certainty of model predictions is rapidly propagated through 
the whole state space, leading to strong generalization and targeted exploration. This in turn
allows the optimal value function to be learned from very few sample transitions: panel (b) shows that
after only 120 transitions (still in the middle of the very first episode) the approximated 
value function already resembles the true one \cite{sutton98introduction}. 
Panel (c) shows the same for a counter-based binary uncertainty; most of the grid cells are 
unvisited and the thus the approximate value function is zero in most parts of the state space.}
\label{fig:5}
\end{figure}

Examining the plots in more detail, we find that, while {\scshape GP-RMAXgrid} is somewhat less sample efficient (explores more),
{\scshape GP-RMAXexp} and {\scshape GP-RMAXnoexp} perform nearly the same. Initially, this appears to be in contrast with 
the whole point of RMAX, which is efficient exploration guided by the uncertainty of the predictions. Here, we believe that this 
behavior can be explained by the good generalization capabilities of GPs. 
Figure~\ref{fig:5} illustrates model learning and certainty propagation with GPs in the mountain car domain 
(predicting acceleration as function of state).
The state of the model-learner is shown for two snapshots: after 40 transitions and after 120 transitions. 
The top row shows the value function that results from applying value iteration with the update modified for 
uncertainty, see Eq.~\eqref{eq:7''}. The bottom row shows the observed samples and the associated certainty 
of the predictions. As expected, certainty is high in regions where data was observed. However, due to the 
generalization of GPs and data-dependent hyperparameter selection, certainty is also high in unexplored regions; 
and in particular it is constant along the $y$-coordinate. To understand this, we have to look at the state 
transition function of the mountain car: acceleration of the car indeed only depends on the position, but not on velocity. 
This shows that certainty estimates of GPs are {\em supervised} and take the properties of the target 
function into account, whereas prior RMAX treatments of uncertainty are {\em unsupervised} and only consider the 
density of samples to decide if a state is ``known''. For comparison, we also show what GP-RMAX with grid-based 
uncertainty would produce in the same situation.

\section{Summary}
We presented an implementation of model-based online reinforcement learning similar to RMAX 
for continuous domains by combining GP-based model learning and value iteration on a grid. 
Doing so, our algorithm separates the problem function approximation in the model-learner
from the problem function approximation/interpolation in the planner. If the transition 
function is easier to learn, i.e., requires only few samples relative to the representation
of the optimal value function, then large savings in sample-complexity can be gained. Related 
model-free methods, such as fitted Q-iteration, can not take advantage of this situation.      
The fundamental limitation of our approach is that it relies on solving the Bellman equation
globally over the state space. Even with more advanced discretization methods, such as adaptive 
grids, or sparse grids, the curse of dimensionality limits the applicability to problems with 
low or moderate dimensionality. Other, more minor limitations, concern the simplifying assumptions
we made: deterministic state transitions and known reward function. However, these are not 
conceptual limitations but rather simplifying assumptions made for the present paper; they
could be easily addressed in future work.

\section*{Acknowledgments}
This work has taken place in the Learning Agents Research
Group (LARG) at the Artificial Intelligence Laboratory, The University
of Texas at Austin.  LARG research is supported in part by grants from
the National Science Foundation (IIS-0917122), ONR (N00014-09-1-0658),
DARPA (FA8650-08-C-7812), and the Federal Highway Administration
(DTFH61-07-H-00030).

\bibliographystyle{plain}

\end{document}